\documentclass[runningheads]{llncs}
\usepackage{graphicx}
\usepackage{hyperref}
\hypersetup{
    colorlinks=true,
    linkcolor=blue,
    citecolor=blue
}

\usepackage{cite}
\begin{document}
\title{Multilingual Evidence Retrieval and Fact Verification to Combat Global Disinformation: The Power of Polyglotism} 
\titlerunning{Multilingual Fact Verification}
\author{Denisa A.O. Roberts}

\authorrunning{D. Roberts}
\institute{AI SpaceTime \\ New York City, NY, USA \\
\email{d.roberts@aispacetime.org}}
\maketitle              % typeset the header of the contribution
\begin{abstract}
 This article investigates multilingual evidence retrieval and fact verification as a step to combat global disinformation, a first effort of this kind, to the best of our knowledge. The goal is building multilingual systems that retrieve in evidence - rich languages to verify claims in evidence - poor languages that are more commonly targeted by disinformation. To this end, our EnmBERT fact verification system shows evidence of transfer learning ability and a 400 example mixed English - Romanian dataset is made available for cross - lingual transfer learning evaluation.
\end{abstract}

\section{Introduction}

 The recent COVID$-19$ pandemic broke down geographical boundaries and led to an \textit{infodemic} of fake news and conspiracy theories \cite{zhou2020recovery}. Evidence based fact verification (English only) has been studied as a weapon against fake news and disinformation \cite{thorne2018fact}. Conspiracy theories and disinformation can propagate from one language to another and some languages are more evidence rich (English). During the US 2020 elections, evidence of online Spanish language disinformation aimed at Latino-American voters was reported \cite{rogers2020}. Polyglotism is not uncommon. According to a 2017 Pew Research study, $91\%$ of European students learn English in school \footnote{\url{https://www.pewresearch.org/fact-tank/2020/04/09/most-european-students-learn-english-in-school/}}. Furthermore, recent machine translation advances are increasingly bringing down language barriers \cite{liu2020multilingual, johnson2017google}. Disinformation can be defined as intentionally misleading information \cite{fallis2015disinformation, fetzer2004disinformation}. The ``good cop'' of the Internet \cite{cohen2018conspiracy}, Wikipedia has become a source of ground truth as seen in the recent literature on evidence-based fact verification. There are more than 6mln English Wikipedia articles \footnote{\url{https://meta.wikimedia.org/wiki/List_of_Wikipedias}} but resources are lower in other language editions, such as Romanian (400K). As a case study we evaluate a claim about Ion Mihai Pacepa, former agent of the Romanian secret police during communism, author of books on disinformation \cite{pacepa2013disinformation, pacepa1987red}. Related conspiracy theories can be found on internet platforms, such as rumors about his death \cite{impact2020}, or Twitter posts in multiple languages, with strong for or against language, such as (English and Portuguese) \footnote{\url{https://twitter.com/MsAmericanPie_/status/1287969874036379649}} or (English and Polish) \footnote{\url{https://twitter.com/hashtag/Pacepa}}. Strong language has been associated with propaganda and fake news \cite{zhou2020survey}. In the following sections we review the relevant literature, present our methodology, experimental results and the case study resolution, and conclude with final notes. We make code, datasets, API, and trained models available \footnote{\url{https://github.com/D-Roberts/multilingual_nli_ECIR2021}}. 
 
\section{Related Work}

The literature review touches on three topics: online disinformation, multilingual NLP and evidence based fact verification. \textbf{Online Disinformation.} Previous disinformation studies focused on election related activity on social media platforms like Twitter, botnet generated hyperpartisan news, 2016 US presidential election \cite{brachten2017strategies, bastos2019brexit, bessi2016social, grinberg2019fake}. To combat online disinformation one must retrieve reliable evidence at scale since fake news tend to be more viral and spread faster \cite{silverman, schroepfer2019creating, zhou2020survey, vosoughi2018spread}. \textbf{Multilingual NLP Advances.} Recent multilingual applications leverage pre-training of massive language models that can be fine-tuned for multiple tasks. For example, the cased multilingual BERT (mBERT) \cite{DBLP:journals/corr/abs-1810-04805},  \footnote{\url{https://github.com/google-research/bert/blob/master/multilingual.md}} is pre-trained on a corpus of the top 104 Wikipedia languages, with 12 layers, 768 hidden units, 12 heads and 110M parameters. Cross-lingual transfer learning has been evaluated for tasks such as: natural language inference \cite{conneau2018xnli, artetxe2019massively}, document classification \cite{schwenk2018corpus}, question answering \cite{clark2020tydi}, fake Indic language tweet detection \cite{kar2020no}. \textbf{English-Only Evidence Retrieval and Fact Verification.} Fact based claim verification is framed as a natural language inference (NLI) task that retrieves its evidence. An annotated dataset was shared \cite{thorne2018fever} and a task \cite{thorne2018fact} was set up to retrieve evidence from Wikipedia documents and predict claim verification status. Recently published SotA results rely on pre-trained BERT flavors or XLNet \cite{yang2019xlnet}. DREAM \cite{zhong2019reasoning}, GEAR \cite{zhou2019gear} and KGAT \cite{liu2020fine} achieved SotA with graphs. Dense Passage Retrieval \cite{karpukhin2020dense} is used in RAG \cite{lewis2020retrieval} in an end-to-end approach for fact verification.

\section{Methodology}

\begin{figure}
\includegraphics[width=\textwidth]{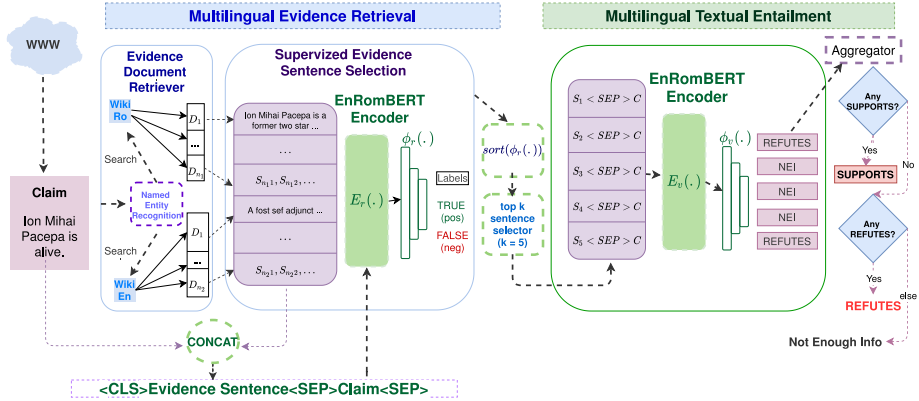}
\caption{Overview of the multilingual evidence retrieval and fact verification system.} \label{fig1}
\end{figure}

The system depicted in Fig. \ref{fig1} is a pipeline with a multilingual evidence retrieval component and a multilingual fact verification component. Based on input claim $c_{l_i}$ in language $l_i$ the system retrieves evidence $E_{l_j}$ from Wikipedia edition in language $l_j$ and supports, refutes or abstains (not enough info). We employ English and Romanian as sample languages.
We use all the annotated $110K$ verifiable claims provided in the initial FEVER task \cite{thorne2018fever} for training the end to end system in Fig. \ref{fig1}.
\textbf{Multilingual Document Retrieval.} To retrieve top Wikipedia $n_l$ documents $D_{c, n_l}$ per claim for each evidence language $l$, we employ an ad-hoc entity linking system \cite{hanselowski2018ukp} based on named entity recognition in \cite{cucerzan2007large}. Entities are parsed from the (English) claim $c$ using the AllenNLP \cite{gardner2018allennlp} constituency parser. We search for the entities and retrieve 7 English \cite{hanselowski2018ukp} and 1 Romanian Wikipedia pages (higher number of Romanian documents did not improve performance) using MediaWiki API \footnote{\url{https://www.mediawiki.org/wiki/API:Main_page}} each. Due to the internationally recognized nature of the claim entities, 144.9K out of 145.5K training claims have Romanian Wikipedia search results. \textbf{Multilingual Sentence Selection.} All sentences $\cup_{n_l}\{S_{D_{c , n_l}}\}$ from each retrieved document are supplied as input to the sentence selection model. We removed diacritics in Romanian sentences \cite{sennrich2016edinburgh} and prepended evidence sentences with the page title to compensate for the missed co-reference pronouns \cite{soleimani2020bert, yoneda2018ucl}. We frame the multilingual sentence selection as a two-way classification task \cite{hanselowski2018ukp, sakata2019faq}. One training example is a pair of an evidence sentence and the claim \cite{zhou2019gear, yoneda2018ucl}. The annotated evidence sentence-claim pairs from FEVER are given the True label. We randomly sample 32 sentences per claim from the retrieved documents as negative sentence-claim pairs (False label). We have 2 flavors of the fine-tuned models: EnmBERT only includes English negative sentences and EnRomBERT includes 5 English and 27 Romanian negative evidence sentences. The architecture includes an mBERT encoder $E_r(\cdot)$ \cite{wolf2019huggingface} \footnote{\url{https://github.com/huggingface/transformers}} and an MLP classification layer $\phi(\cdot)$. During training, all the parameters are fine-tuned and the MLP weights are trained from scratch. The encoded first $<CLS>$ token, is supplied to the MLP classification layer. For each claim, the system outputs all the evidence sentence-claim pairs ranked in the order of the predicted probability of success  $P(\mathbf{y}=1|\mathbf{x}) = \phi(E_{r}(\mathbf{x}))$ (pointwise ranking \cite{cao2007learning}). \textbf{Multilingual Fact Verification.} The fact verification step (NLI) training takes as input the 110K training claims paired with each of the 5 selected evidence sentences (English only for EnmBERT or En and Ro for EnRomBERT), and fine-tunes the three-way classification of pairs using the architecture in Fig. \ref{fig1}). We aggregate the predictions made for each of the 5 evidence sentence-claim pairs based on logic rules \cite{malon2018team}(see Fig. \ref{fig1}) to get one prediction per claim. Training of both sentence selection and fact verification models employed the Adam optimizer \cite{kingma2014adam}, batch size of 32, learning rate of $2e-5$, cross-entropy loss, and 1 and 2 epochs of training, respectively. \textbf{Alternative Conceptual End-to-End Multilingual Retrieve-Verify System.} The entity linking approach to document retrieval makes strong assumptions about the presence of named entities in the claim. Furthermore, the employed constituency parser \cite{gardner2018allennlp} assumes that claims are in English. To tackle these limitations, we propose a conceptual end-to-end multilingual evidence retrieval and fact verification approach inspired by the English-only RAG ~\cite{lewis2020retrieval}. The system automatically retrieves relevant evidence passages in language $l_j$ from a multilingual corpus corresponding to a claim in language $l_i$. In Fig. \ref{fig1}, the 2-step multilingual evidence retrieval is replaced with a multilingual version of dense passage retrieval (DPR) \cite{karpukhin2020dense} with mBERT backbone. The retrieved documents form a latent probability distribution. The fact verification step conditions on the claim $x_{l_i}$ and the latent retrieved documents $z$ to generate the label $y$, $P(y|x_{l_i}) = \sum_{z \in D_{top-k, l_j}} p(z|x_{l_i})p(y|x_{l_i}, z)$. The multilingual retrieve-verify system is jointly trained and the only supervision is at the fact verification level. We leave this promising avenue for future experimental evaluation.

\section{Experimental Results}

\begin{figure}
\includegraphics[width=\textwidth, height=5cm]{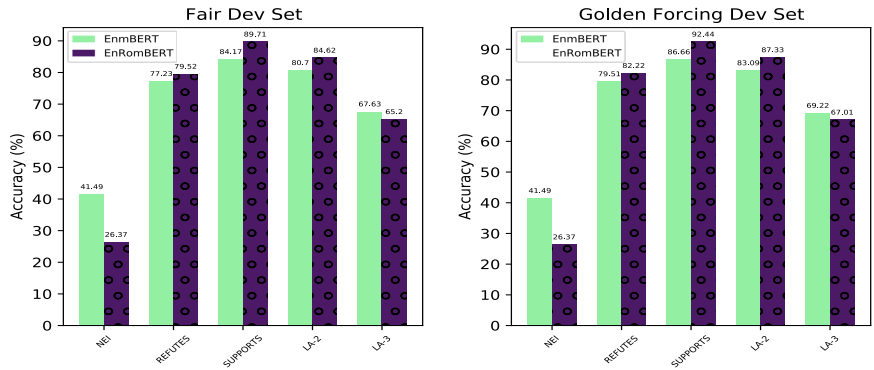}
\caption{Error analysis per class. `LA-2' is Accuracy for `Supports' $\&$ `Refutes' Claims}\label{fig2}
\end{figure}

 In the absence of equivalent end-to-end multilingual fact verification baselines, we compare performance to English-only systems using the official FEVER scores \footnote{\url{https://github.com/sheffieldnlp/fever-scorer}} on the original FEVER datasets \cite{thorne2018fever}. Furthermore, the goal of this work is to use multilingual systems trained in evidence rich languages to combat disinformation in evidence poor languages. To this end we evaluate the transfer learning ability of the trained verification models on an English-Romanian translated dataset. We translated 10 supported and 10 refuted claims (from the FEVER developmental set) together with 5 evidence sentences each (retrieved by the EnmBERT system) and combined in a mix and match development set of 400 examples. \textbf{Calibration results on FEVER development and test sets.} In Table \ref{tab1} and Fig. \ref{fig2} we compare EnmBERT and EnRomBERT verification accuracy (LA-3) and evidence recall on the fair FEVER development (dev) set, the test set and on a golden-forcing dev set. The fair dev set includes all the claims in the original FEVER dev set and all the sentences from the retrieved documents (English and/or Romanian). The golden forcing dev set forces all ground truth evidence into the sentence selection step input, effectively giving perfect document retrieval recall \cite{liu2020fine}. On the fair dev set, the EnmBERT system reaches within $5\%$ accuracy of English-only BERT-based systems such as \cite{soleimani2020bert} (LA-3 of 67.63\%). We also reach within $5\%$ evidence recall (Table \ref{tab1} $88.60\%$) as compared to English-only KGAT \cite{liu2020fine} and better than \cite{soleimani2020bert}. Note that any of the available English-only systems with BERT backbone such as KGAT \cite{liu2020fine} and GEAR \cite{zhou2019gear} can be employed with an mBERT (or another multilingual pre-trained) backbone to lift the multilingual system performance. 
\begin{table}
\centering
\caption{Calibration of models evaluation using the official FEVER scores $\%$ in \cite{thorne2018fever}.}\label{tab1}
\begin{tabular}{l|l | l l l l l}
\hline
 Dataset   & Model               &   Prec@5 &   Rec@5  &   FEVER &   LA-3 Acc \\
\hline
 Fair-Dev     & EnmBERT-EnmBERT &    25.54 &   $\mathbf{88.60}$  &   64.62 & $\mathbf{67.63}$ \\
 Fair-Dev     & EnRomBERT-EnRomBERT &    25.20 &   88.03  &   61.16 & 65.20 \\
\hline
 Test     & EnmBERT-EnmBERT &    25.27 &   87.38  &   62.30 & 65.26 \\
 Test     & EnRomBERT-EnRomBERT &   24.91  &   86.80  &   58.78 &63.18  \\
 \hline
\end{tabular}
\end{table}
To better understand strengths and weaknesses of the system performance and the impact of including Romanian evidence in training EnRomBERT, we present a per class analysis in Fig. 2 \ref{fig2}. We also calculate accuracy scores for only `SUPPORTS' and `REFUTES' claims (FEVER-2). The English-only SotA label accuracy (LA-2) on FEVER-2 is currently given in RAG \cite{lewis2020retrieval} at $89.5\%$ on the fair dev set and our EnRomBERT system reaches within $5\%$. We postulate that the noise from including Romanian sentences in training improves the FEVER-2 score (see Fig. \ref{fig2}), EnRomBERT coming within $5\%$ of \cite{thorne2020avoiding} English-only FEVER-2 SotA of $92.2\%$ on the golden-forcing dev set. In the per-class analysis, on 'SUPPORTS' and 'REFUTES' classes in Fig. \ref{fig2}, EnRomBERT outperforms EnmBERT on both fair and golden-forcing dev sets. To boost the NEI class performance, future research may evaluate the inclusion of all claims, including NEI, in training. Furthermore, retrieval in multiple languages may alleviate the absence of relevant evidence for NEI claims.\textbf{Transfer Learning Performance} Table ~\ref{tab2} shows EnmBERT and EnRomBERT transfer learning ability evaluated directly in the fact verification step using the previously retrieved and manually translated 400 mixed claim-evidence pairs. We report the classification accuracy on all 400 mixed examples, and separately for En-En (English evidence and English claims), En-Ro, Ro-En and Ro-Ro pairs. EnmBERT's zero-shot accuracy on Ro-Ro is $85\%$ as compared to $95\%$ for En-En, better than EnRomBERT's. EnmBERT outperforms EnRomBERT as well for Ro-En and En-Ro pairs. We recall that Romanian evidence sentences were only included in EnRomBERT training as negative evidence in the sentence retrieval step. If selected in the top 5 evidence sentences, Romanian sentences were given the NEI label in the fact verification step. Hence, EnRomBERT likely learned that Romanian evidence sentences are NEI, which led to a model bias against Romanian evidence. \textbf{Disinformation Case Study} We employ EnmBERT to evaluate the claim ``Ion Mihai Pacepa, the former Securitate general, is alive''. The document retriever retrieves Wikipedia documents in English, Romanian and Portuguese. Page summaries are supplied to the EnmBERT sentence selector, which selects top 5 evidence sentences (1XEn, 2XRo, 2XPt). Based on the retrieved evidence, the EnmBERT fact verification module predicts `SUPPORTS' status for the claim. For illustration purposes, the system is exposed as an API \footnote{\url{https://github.com/D-Roberts/multilingual_nli_ECIR2021}}. 
\begin{table}
\centering
\caption{Fact Verification Accuracy ($\%$) for Translated Parallel Claim - Evidence Sentences.}\label{tab2}
\begin{tabular}{l|l l l l l}
\hline
 Model        &   Mixed &   En-En &   En-Ro &   Ro-En &   Ro-Ro  \\
\hline
 EnmBERT &      95.00 &     95.00 &     50.00 &     65.00 &     85.00 \\

 EnRomBERT &      95.00 &     95.00 &     25.00 &      0.00 &     50.00 \\
\hline
\end{tabular}
\end{table}

 \section{Final Notes}
 
In this article we present a first approach to building multilingual evidence retrieval and fact verification systems to combat global disinformation. Evidence poor languages may be at increased risk of online disinformation and multilingual systems built upon evidence rich languages in the context of polyglotism can be an effective weapon. To this end, our trained EnmBERT system shows cross-lingual transfer learning ability for the fact verification step on the original FEVER-related claims. This work opens future lines of research into end-to-end multilingual retrieve-verify systems for disinformation suspect claims, in multiple languages, with multiple reliable evidence retrieval sources available in addition to Wikipedia.

\bibliographystyle{splncs04}
\bibliography{mybib}

\end{document}